\documentclass[lettersize,journal]{IEEEtran}
\usepackage{amsmath,amsfonts}
\usepackage{algorithmic}
\usepackage{algorithm}
\usepackage{array}
\usepackage[caption=false,font=normalsize,labelfont=sf,textfont=sf]{subfig}
\usepackage{textcomp}
\usepackage{stfloats}
\usepackage{url}
\usepackage{verbatim}
\usepackage{graphicx}
\usepackage{cite}
\begin{document}

\title{Attribute Inference from Interactive Targeted Ads}

\author{Peihao Li}

\maketitle

\begin{abstract}
Targeted advertising systems can pair audiences selected by advertisers with ad units that expose visible user actions. When an interaction remains linked to the campaign that elicited it, the advertiser may receive an observation tied to a user rather than only an aggregate report. We model that channel as a noisy oracle for attribute inference. The model separates targeting predicates, exposure, interaction, and disclosure. These boundaries capture the gap between eligibility and delivery, and the gap between interaction and advertiser visibility.

We build a reproducible benchmark using synthetic populations calibrated with public data, each with known sensitive labels. A generated campaign semantics layer provides topic variants and response priors. The simulator generates the ground truth, event traces, disclosed observations, and metrics. The evaluation compares Bayesian, supervised, positive and unlabeled, and adaptive attacks under common campaign and disclosure definitions.

The final evaluation uses four topic variants, seven simulator seeds, and two interaction settings. Repeated campaigns with identity exposure produce measurable but bounded inference signal. At $160$ campaigns, Bayesian and supervised attacks reach about $0.64$ AUC in the main setting and about $0.65$ AUC in the higher interaction setting. Disclosure policy is the strongest control. Aggregate reporting removes the evaluated oracle input tied to users. Type filtering and randomized disclosure reduce the released signal. The result is a model, artifact, and defense evaluation method for privacy in interactive targeted advertising. The code is available at \url{https://github.com/P-HOW/Interactive-Ad-Oracle}.
\end{abstract}


\section{Introduction}
\IEEEPARstart{S}{ocial} platforms now combine targeted advertising with ad units that invite visible user action. An advertiser defines an audience and campaign criteria through platform tools. The advertiser then observes delivery reports. In some platform designs, the advertiser can also inspect user-facing traces created by the ad. Prior work has shown that targeting and reporting interfaces can leak information through audience construction and reach estimates \cite{Korolova2011PrivacyViolationsUsingMicrotargeted,Venkatadri2018PrivacyRisksFacebooks}. Studies of ad explanations show that transparency interfaces can reveal partial views of the targeting process \cite{Andreou2018InvestigatingAdTransparencyMechanisms}. Interactive ads add a signal tied to users. A platform may expose a reaction or comment as an identifiable engagement record to the advertiser \cite{Kieserman2026RecklessDesignsBrokenPromises}. When this signal is tied to a targeted campaign, the advertising system can produce observations tied to users rather than aggregate campaign statistics alone. The privacy question is whether interactions that expose identity on targeted ads can act as noisy oracles for sensitive attribute inference.

The attack follows from the interface. An advertiser can treat a campaign as a query against the platform's targeting system. The targeting predicate defines a subset of eligible users or user profiles. One predicate may match a custom audience. Another may match an interest group inferred by the platform or a demographic range. If an identifiable user interacts with the ad, the advertiser learns an observation about that user under a predicate chosen before the campaign ran. The observation is noisy because targeting eligibility is distinct from ad delivery, while delivery does not imply attention or interaction. The noise changes the inference problem rather than removing it. Repeated campaigns create an observation history in which each visible interaction supplies evidence about a user's relationship to a known predicate. Across enough predicates, that history can support probabilistic inference about attributes that were never directly disclosed to the advertiser. This composition of predicates chosen by the advertiser, interactions that expose identity, and repeated campaigns is the systems privacy problem studied here \cite{Korolova2011PrivacyViolationsUsingMicrotargeted,Kieserman2026RecklessDesignsBrokenPromises}.

The work builds on prior research in advertising privacy, transparency, delivery, tracking, and inference. Studies of microtargeting and advertiser interfaces have shown that ad systems can leak information through platform feedback and audience tools \cite{Korolova2011PrivacyViolationsUsingMicrotargeted,Venkatadri2018PrivacyRisksFacebooks}. Work on ad transparency has examined how users and auditors observe targeting explanations \cite{Andreou2018InvestigatingAdTransparencyMechanisms}. Work on ad delivery has shown that platform optimization can produce discriminatory outcomes even when the stated targeting choices appear neutral \cite{Ali2019DiscriminationthroughOptimizationHow}. Measurement studies of web tracking have documented infrastructure that links browsing activity, identifiers, and behavioral profiling across sites \cite{Englehardt2016OnlineTracking1million}. Privacy attacks in machine learning provide a related method base for inferring hidden properties from observed system outputs. This base includes membership inference, model inversion, and attribute inference \cite{Shokri2017MembershipInferenceAttacksagainst,Jayaraman2022AreAttributeInferenceAttacks,Rigaki2023SurveyPrivacyAttacksMachine}. The missing problem is the interaction between targeting predicates chosen by advertisers and ad interactions that expose identity. Existing work has not provided a systematic account of this interaction as an oracle, or a benchmark for attack and defense evaluation in this setting.

We organize the paper around four research questions. \textbf{RQ1} asks whether interactions that expose identity enable sensitive attribute inference when the advertiser knows the predicates it selected and observes the users who interact with its ads. \textbf{RQ2} asks how risk changes with campaign count, interaction rate, targeting granularity, minimum audience constraints, and disclosure policy. \textbf{RQ3} asks whether adaptive campaign selection reduces the campaigns or cost needed to reach a target confidence level. \textbf{RQ4} asks which platform side disclosure defenses reduce the inference channel tied to users while preserving ordinary campaign reporting.

We answer these questions by introducing an oracle abstraction for interactive targeted ads. The abstraction separates the campaign predicate chosen by the advertiser from the latent user attribute being inferred. It also separates both quantities from the noisy interaction process that creates visible evidence. We instantiate this abstraction in a campaign predicate model with audience constraints and disclosure rules. The attack section instantiates the inference rule and the campaign selection policy. The benchmark section instantiates the user population, campaign library, event process, disclosure policies, and calibration splits. The evaluation then answers RQ1 through RQ4 under common event traces and shared metrics.

The evaluation isolates the inference channel under controlled disclosure policies. Synthetic users calibrated from public data provide known ground truth for sensitive attributes. They support parameter sweeps and repeatable attack comparisons without collecting sensitive labels from real users without consent. This benchmark design follows the role of controlled datasets in privacy inference evaluation \cite{Jayaraman2022AreAttributeInferenceAttacks}. It also follows recent work on synthetic data for personal attribute inference \cite{Yukhymenko2024SyntheticDatasetPersonalAttribute}. Controlled platform validation, where used, checks whether the assumed disclosure channels exist under consented account conditions. The defense evaluation includes group size constraints and aggregate reporting. It also studies noisy release mechanisms. These choices draw on k-anonymity and differential privacy as formal privacy tools for reducing individual disclosure from released data \cite{Sweeney2002kAnonymityModelProtecting,Dwork2006CalibratingNoiseSensitivityPrivate}.

The contributions are:
\begin{itemize}
\item We formalize interactive targeted ads as noisy attribute inference oracles under disclosure policies with visible identity. This answers RQ1 by defining the observation channel tied to users and the conditions under which campaign predicates become inference evidence.
\item We instantiate the oracle with Bayesian, supervised, positive and unlabeled, and adaptive attacks. This answers RQ3 by separating the inference rule $f$ from the campaign selection policy $\pi$ and by defining confidence and cost metrics for repeated campaigns.
\item We build a reproducible benchmark with synthetic users and calibration from public data for evaluating RQ1 and RQ2. The benchmark supports sweeps over campaign count, interaction rate, predicate granularity, minimum audience constraints, and disclosure rules.
\item We evaluate disclosure defenses for RQ4. The defense space covers type filtering, randomized identity disclosure, thresholding, and aggregate reporting as transformations of the disclosure map $D$.
\end{itemize}

\section{Background and Motivation}

\begin{figure*}[t]
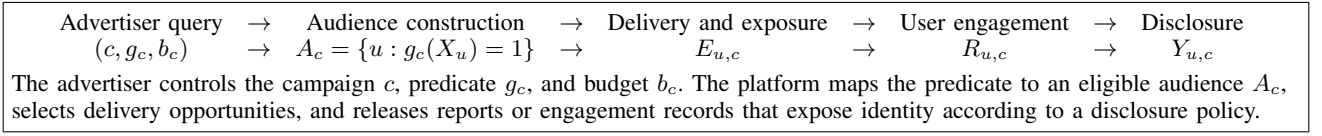

\centering
\fbox{%
\begin{minipage}{0.94\textwidth}
\centering
\small
\begin{tabular}{c@{\quad$\rightarrow$\quad}c@{\quad$\rightarrow$\quad}c@{\quad$\rightarrow$\quad}c@{\quad$\rightarrow$\quad}c}
Advertiser query & Audience construction & Delivery and exposure & User engagement & Disclosure \\
$(c,g_c,b_c)$ & $A_c=\{u:g_c(X_u)=1\}$ & $E_{u,c}$ & $R_{u,c}$ & $Y_{u,c}$
\end{tabular}
\vspace{4pt}

\raggedright
The advertiser controls the campaign $c$, predicate $g_c$, and budget $b_c$. The platform maps the predicate to an eligible audience $A_c$, selects delivery opportunities, and releases reports or engagement records that expose identity according to a disclosure policy.
\end{minipage}}
\caption{Pipeline view of an interactive targeted ad as a query surface. The formal model in Section~\ref{sec:oracle-model} assigns probability models to delivery, interaction, and disclosure.}
\label{fig:ad-oracle-pipeline}
\end{figure*}

The oracle abstraction in the introduction depends on separating the stages of a targeted advertising system. A campaign records the advertiser's objective and budget, together with the ad creative. The targeting predicate is the advertiser's chosen condition over platform data. It may refer to platform attributes, inferred interests, advertiser supplied lists, or audience planning categories \cite{Beauvisage2023Howonlineadvertisingtargets}. Audience construction applies that predicate before the delivery system selects impression opportunities. Exposure records that an ad was shown to a user. Engagement records a user action on the ad surface. Reporting releases campaign statistics. Disclosure policy determines whether the advertiser receives aggregate measurements or traces tied to users. Figure~\ref{fig:ad-oracle-pipeline} summarizes the pipeline and the variables used throughout the paper.

Prior audits show why these stages cannot be collapsed into a single report. Microtargeting and PII based targeting can leak information when small predicates reveal whether a person satisfies an advertiser selected condition \cite{Korolova2011PrivacyViolationsUsingMicrotargeted,Venkatadri2018PrivacyRisksFacebooks}. Delivery is also a separate source of risk. Platform optimization can change which eligible users receive ads after the predicate is set, producing privacy and fairness concerns even under neutral stated criteria \cite{Ali2019DiscriminationthroughOptimizationHow,Br2024Systematicdiscrepanciesdeliverypolitical}. We treat the advertising interface as a query surface. The advertiser chooses predicates first. The advertiser later observes delivery outcomes and engagement events. Campaign reports and disclosure rules determine how those observations are released. This separation follows audit work that studies targeted advertising through interface controls and observed system behavior \cite{Lam2023SociotechnicalAuditsBroadeningAlgorithm}.

Transparency and disclosure interfaces expose different facts to different actors. User facing ad explanations give a partial account of why a person saw an ad, often simplifying the path from targeting to delivery \cite{Andreou2018InvestigatingAdTransparencyMechanisms}. Public ad libraries expose archived ads and selected metadata for outside observers. Security analyses show that these archives can suffer from evasion and completeness failures \cite{Edelson2020SecurityAnalysisFacebookAd}. Advertiser analytics reports expose aggregate campaign performance, including impressions or reach. Interaction disclosure interfaces differ because they can reveal an identifiable engagement trace to the advertiser who ran the campaign \cite{Kieserman2026RecklessDesignsBrokenPromises}.

The privacy channel studied here is an interaction that exposes identity. Social media studies treat comments and likes as distinct engagement modes. Reactions and shares are also observable response types on many social surfaces \cite{Tenenboim2022CommentsSharesorLikes}. An aggregate engagement count tells the advertiser that some users responded. A record that exposes identity tells the advertiser which user responded. When that record is tied to a campaign predicate chosen before delivery, the interaction becomes evidence about the user's relationship to that predicate. The evidence remains noisy because eligibility does not imply delivery. A delivered impression also does not imply exposure, attention, or a decision to interact. The noise changes inference accuracy, but it does not remove the channel.

A single visible engagement supplies evidence under one advertiser chosen predicate. The systems risk arises when the advertiser repeats the process. Each campaign supplies a chosen condition. Each engagement that exposes identity supplies an observation under that condition. Across campaigns, the advertiser obtains a history $H_u^{(T)}$ linking predicates to responses by user after $T$ campaigns. Such histories can support attribute inference because targeting predicates partition users in ways that may correlate with latent traits. Prior work on microtargeting and nanotargeting shows that advertising predicates can become highly identifying when they isolate small groups or combine several attributes \cite{Korolova2011PrivacyViolationsUsingMicrotargeted,GonzlezCabaas2021UniqueFacebookFormulationEvidence}.

The inference problem is related to privacy attacks on observed system outputs, but the observable object is different. Attribute inference studies whether partial knowledge and observed outputs can reveal missing sensitive values. It also stresses comparison with imputation baselines, which is relevant for evaluating claims in this setting \cite{Jayaraman2022AreAttributeInferenceAttacks}. In an interactive ad oracle, the output is not a classifier score. It is an engagement record that exposes identity after the platform applies targeting, delivery, and disclosure rules. The attacker still learns from chosen inputs and observed outputs, but the input is a campaign predicate and the output is a disclosed interaction trace.

Delivery behavior remains part of the oracle model because the delivered audience can differ from the population implied by the predicate alone \cite{Ali2019DiscriminationthroughOptimizationHow,Br2024Systematicdiscrepanciesdeliverypolitical}. Tracking and profiling can also affect audience construction by changing correlations among predicates and users \cite{Englehardt2016OnlineTracking1million}. The model keeps eligibility separate from delivery. It also separates engagement, reporting, and disclosure. This structure lets the attack evaluation vary each component without treating any one platform report as ground truth.

A controlled benchmark for this setting needs known attributes, repeatable campaign predicates, and disclosure rules that can be varied across runs. Synthetic or public calibrated users provide ground truth for sensitive attributes while avoiding live experiments that collect sensitive labels from real platform users \cite{Yukhymenko2024SyntheticDatasetPersonalAttribute,Jayaraman2022AreAttributeInferenceAttacks}. The same population can support sweeps over campaign count, interaction rate, targeting granularity, and audience size constraints. Public calibration gives the synthetic population realistic attribute distributions. Synthetic records keep the evaluation repeatable while allowing the benchmark to vary disclosure policy.

Defenses map to the same interface boundaries as the oracle. Predicate constraints can remove sensitive targeting categories or impose minimum audience sizes before a campaign can run \cite{Korolova2011PrivacyViolationsUsingMicrotargeted,Venkatadri2018PrivacyRisksFacebooks}. External evaluations of mitigation measures show that platform changes can be studied through observed delivery outcomes, which motivates defense evaluation at both the targeting and delivery layers \cite{Imana2025ExternalEvaluationDiscriminationMitigation}. Disclosure limits can hide identities in engagement records. Reporting defenses can replace traces tied to users with aggregate reporting. Group thresholds and k-anonymity formalize the rule that released records should not single out individuals \cite{Sweeney2002kAnonymityModelProtecting}. Randomized response motivates perturbing categorical reports before release \cite{Warner1965RandomizedResponseSurveyTechnique}. Differential privacy gives a formal basis for calibrated noisy release of aggregate statistics \cite{Dwork2014AlgorithmicFoundationsDifferentialPrivacy}.

\section{Threat Model}

The threat model fixes the same pipeline at a lower level of abstraction. Let $U$ denote users, $\mathcal{C}$ denote campaigns, $g_c$ denote the targeting predicate selected for campaign $c$, $S_u$ denote the sensitive benchmark attribute associated with user $u$, and $H_u^{(T)}$ denote the observation history for user $u$ after $T$ campaigns. The attacker is an advertiser operating through the ordinary campaign interface. Its inputs are campaigns and predicates. Its outputs are the records released by reporting and disclosure rules.

\begin{table*}[t]
\centering
\caption{Threat model summary.}
\label{tab:threat-model}
\renewcommand{\arraystretch}{1.15}
\begin{tabular}{p{0.20\textwidth}p{0.74\textwidth}}
\hline
Component & Model used in the paper \\
\hline
Attacker & An advertiser that creates campaigns, selects platform accepted predicates $g_c$, sets budgets or campaign limits, observes campaign reports, and inspects interactions that expose identity when the disclosure policy exposes them. \\
Targets & Users whose identities become visible through one or more campaign interactions, plus controlled or consented accounts in validation experiments. \\
Hidden attribute & A benchmark label $S_u$ such as health interest, political interest, financial stress, or parent status. The platform releases campaign observations rather than $S_u$. \\
Observation history & A sequence $H_u^{(T)}$ of campaign metadata, predicates, timing, aggregate reports, and disclosed interactions linked to user $u$ after $T$ campaigns. \\
Constraints & Minimum audience sizes, predicate-combination limits, sensitive-category rules, budget limits, delivery rules, approval checks, and disclosure rules. \\
Excluded capabilities & Platform compromise, account takeover, malware, private API abuse, unauthorized collection of private user records, and unparameterized linkage through external data brokers. \\
\hline
\end{tabular}
\end{table*}

The attacker knows the predicates it selected and the timing of its own campaigns. In the fixed setting, it chooses a campaign set before seeing any observations. In the adaptive setting, it uses earlier reports or disclosed interactions to choose later campaigns \cite{Lam2023SociotechnicalAuditsBroadeningAlgorithm}. Adaptation changes the query strategy, while the action space remains the platform campaign interface. The attacker treats each campaign as a chosen query. Its result reflects targeting and delivery effects. It also reflects user interaction behavior and disclosure policy. This view follows audit methods that study advertising systems by varying interface inputs and observing released outputs \cite{Lam2023SociotechnicalAuditsBroadeningAlgorithm}.

The target is a user whose identity becomes visible to the advertiser through one or more campaign interactions. In validation experiments, a target can also be a controlled or consented account whose attributes are known to the experimenter. The sensitive attribute is a private label represented in the benchmark. Examples include health interest and political interest. Other benchmark labels can encode financial stress or parent status. The attack objective is to estimate $S_u$ from the target's observation history $H_u^{(T)}$ rather than from an explicit release of the sensitive label \cite{Jayaraman2022AreAttributeInferenceAttacks}.

The model distinguishes direct sensitive predicates from proxy predicates. A direct sensitive predicate is closely tied to the label, when the platform permits such a criterion under its rules. A proxy predicate is a nonsensitive feature with statistical dependence on the label. The attack can use either type when the predicate is permitted and the campaign satisfies the platform constraints. Attribute inference work frames this task as estimating hidden values from observed records and auxiliary structure, rather than from an explicit release of the target attribute \cite{Shokri2017MembershipInferenceAttacksagainst,Jayaraman2022AreAttributeInferenceAttacks}.

The core observation is an interaction that exposes identity linked to a campaign. The record binds a visible user identity to a campaign chosen by the advertiser and to the disclosure policy that made the engagement inspectable. The exact interaction type depends on platform design. The model covers visible comments and reactions when the interface exposes them to the advertiser \cite{Kieserman2026RecklessDesignsBrokenPromises}. It also covers other visible engagement traces under the same disclosure rule. The observation history for a target is the sequence of disclosed interactions, paired with the predicates and timing of the campaigns that produced them \cite{Kieserman2026RecklessDesignsBrokenPromises}.

Aggregate reports are also observations. Reach and impression measurements give campaign level delivery signals \cite{Venkatadri2018PrivacyRisksFacebooks}. Engagement totals give campaign level response signals. These signals can help the attacker choose later campaigns, estimate base rates, or decide whether to continue a campaign sequence. The available records consist of disclosed interactions, aggregate reports, and campaign metadata selected by the advertiser. The hidden part includes eligible users who receive no delivered ad, exposed users who take no visible action, and users whose interactions are hidden by the disclosure policy. This separation is needed because campaign reports and disclosures tied to users expose different parts of the advertising pipeline \cite{Lam2023SociotechnicalAuditsBroadeningAlgorithm}. Aggregate releases are relevant to privacy because repeated aggregate observations can support inference under realistic attacker knowledge \cite{Guan2024ZeroAuxiliaryKnowledgeMembership}.

Platform constraints are part of the threat model. Campaigns must satisfy minimum audience sizes, predicate-combination limits, sensitive-category rules, and budget limits. Delivery rules, approval checks, and disclosure rules further restrict the query surface \cite{Andreou2018InvestigatingAdTransparencyMechanisms}. The adaptive attacker chooses legal campaigns within those constraints \cite{Korolova2011PrivacyViolationsUsingMicrotargeted,Venkatadri2018PrivacyRisksFacebooks}. Delivery noise is included because eligible users and delivered impressions can differ after platform allocation. Constraints affect attack cost and inference accuracy. Prior studies of microtargeting and advertising interfaces motivate treating audience-size thresholds and predicate restrictions as relevant to privacy controls \cite{Korolova2011PrivacyViolationsUsingMicrotargeted,Venkatadri2018PrivacyRisksFacebooks}.

The base model consists of campaign creation, targeting predicate selection, budget selection, aggregate reporting, and interaction disclosure through advertising and reporting interfaces provided by the platform \cite{Lam2023SociotechnicalAuditsBroadeningAlgorithm}. It covers repeated campaigns and adaptive selection over the same interface. The base model fixes the advertiser side oracle to records exposed by campaigns, reports, and disclosure policy. Platform compromise, account takeover, malware, private API abuse, unauthorized collection of private user records, and unparameterized linkage through external data brokers belong to other attacker models. Parameterized auxiliary information, browser privacy, and tracking strength controls can be added without changing the oracle interface because they change predicate precision and correlation structure rather than the disclosure map.

The benchmark uses synthetic users or users calibrated from public data so that sensitive labels are known without collecting sensitive labels from real platform users outside the study. Synthetic personal attribute benchmarks support repeatable inference evaluation with controlled ground truth \cite{Yukhymenko2024SyntheticDatasetPersonalAttribute}. The evaluation treats attribute labels as benchmark variables, not as labels gathered from nonparticipants. Public data calibration sets distributions and correlations. Synthetic users supply the records needed for attack comparison \cite{Yukhymenko2024SyntheticDatasetPersonalAttribute,Jayaraman2022AreAttributeInferenceAttacks}.

Controlled validation checks disclosure behavior with controlled or consented accounts. The validation question is whether a platform interface exposes the interaction record described by the threat model. Attribute inference is evaluated on benchmark users and controlled accounts. Retained evidence records aggregate interface behavior or redacted observations, while sensitive labels and nonparticipant identities remain outside the benchmark records. Group thresholds and calibrated noise provide the basis for later reporting defenses \cite{Sweeney2002kAnonymityModelProtecting,Dwork2006CalibratingNoiseSensitivityPrivate}. These capabilities, constraints, and observations define the formal oracle model used by the attacks and defenses.

\section{System and Oracle Model}
\label{sec:oracle-model}

This section formalizes the oracle used by the attacks and the benchmark. An interactive targeted ad is a noisy query whose output is determined by eligibility, exposure, interaction, and disclosure.

\begin{table}[t]
\centering
\small
\begin{tabular}{@{}p{0.21\linewidth}p{0.67\linewidth}@{}}
\hline
Symbol & Meaning \\
\hline
$U$ & finite user population \\
$u$ & one user in $U$ \\
$X_u$ & platform side features for user $u$ \\
$S_u$ & binary sensitive benchmark attribute \\
$\mathcal{C}$ & campaign space \\
$\mathcal{C}_{\mathrm{legal}}$ & legal campaign space after platform constraints \\
$c=(g_c,a_c,b_c)$ & campaign, creative or topic, and budget or normalized cost \\
$A_c$ & eligible audience induced by $g_c$ \\
$E_{u,c}$ & exposure event \\
$R_{u,c}$ & interaction event \\
$T_{u,c}$ & interaction type \\
$D$ & disclosure policy \\
$Y_{u,c}$ & visible to advertisers observation tied to a user \\
$Y_c$ & aggregate campaign observation \\
$H_u^{(T)}$ & history indexed by user after $T$ campaigns \\
$G^{(T)}$ & aggregate history after $T$ campaigns \\
$f$ & inference rule \\
$\pi$ & campaign selection policy \\
\hline
\end{tabular}
\caption{Notation used in the oracle model.}
\label{tab:oracle-notation}
\end{table}

\paragraph{Users and campaigns.}
Let $U=\{u_1,\ldots,u_N\}$ be a finite population. Each user $u\in U$ has features $X_u\in\mathcal{X}$ and a binary benchmark attribute $S_u\in\{0,1\}$. The advertiser does not observe $S_u$ at test time. A campaign is
\begin{equation}
c=(g_c,a_c,b_c),
\end{equation}
where $g_c:\mathcal{X}\rightarrow\{0,1\}$ is the targeting predicate, $a_c$ is the creative or topic, and $b_c\ge 0$ is budget or normalized cost. The eligible audience is
\begin{equation}
A_c=\{u\in U:g_c(X_u)=1\}.
\end{equation}
The legal campaign space is
\begin{equation}
\mathcal{C}_{\mathrm{legal}}
=
\{c\in\mathcal{C}: |A_c|\ge m,\ b_c\le B_{\max},\ g_c\in\mathcal{G}_{\mathrm{allowed}}\}.
\end{equation}
The threshold $m$, budget cap $B_{\max}$, and allowed predicate set $\mathcal{G}_{\mathrm{allowed}}$ encode the platform constraints used in RQ2 and RQ4.

\paragraph{Exposure and interaction.}
Eligibility changes the exposure distribution but does not determine exposure. For each user and campaign,
\begin{equation}
E_{u,c}\sim \mathrm{Bernoulli}(p^E_{u,c}),
\qquad
p^E_{u,c}=p_E(u,c;g_c,a_c,b_c,\theta_E).
\end{equation}
The model assumes targeting monotonicity:
\begin{equation}
p_E(u,c\mid g_c(X_u)=1)\ge p_E(u,c\mid g_c(X_u)=0).
\end{equation}
Interaction is conditional on exposure:
\begin{equation}
\Pr(R_{u,c}=1\mid E_{u,c}=0)=0.
\end{equation}
\begin{equation}
\Pr(R_{u,c}=1\mid E_{u,c}=1)
=
p_R(u,c;X_u,S_u,a_c,\theta_R).
\end{equation}
The interaction type $T_{u,c}\in\mathcal{T}$ records the visible action category when an interaction occurs. The benchmark instantiates $p_E$ and $p_R$ through the campaign and interaction parameters evaluated in RQ2.

\paragraph{Disclosure and histories.}
Let $\bot$ denote no released record tied to a user. A disclosure policy maps event traces to observations visible to advertisers. Disclosure with visible identity releases a record tied to a user for interactions:
\begin{equation}
Y_{u,c}^{\mathrm{id}}=
\begin{cases}
(u,c,T_{u,c}), & R_{u,c}=1,\\
\bot, & R_{u,c}=0.
\end{cases}
\end{equation}
Type filtering releases the same record only for shown action types $\mathcal{T}_{\mathrm{show}}\subseteq\mathcal{T}$:
\begin{equation}
Y_{u,c}^{\mathrm{type}}=
\begin{cases}
(u,c,T_{u,c}), & R_{u,c}=1\ \text{and}\ T_{u,c}\in\mathcal{T}_{\mathrm{show}},\\
\bot, & \text{otherwise}.
\end{cases}
\end{equation}
Randomized identity disclosure samples otherwise visible records with probability $\rho$. Threshold disclosure releases records that expose identity only when $\sum_{v\in U}R_{v,c}\ge k$. Aggregate reporting releases
\begin{equation}
Y_c^{\mathrm{agg}}=\sum_{u\in U}R_{u,c}
\end{equation}
and no $Y_{u,c}$ indexed by user.

For reporting tied to users, the target history after $T$ campaigns is
\begin{equation}
H_u^{(T)}=
\left((c_t,g_{c_t},a_{c_t},b_{c_t},Y_{u,c_t})\right)_{t=1}^{T}.
\end{equation}
For aggregate reporting, the aggregate history is
\begin{equation}
G^{(T)}=
\left((c_t,g_{c_t},a_{c_t},b_{c_t},Y_{c_t})\right)_{t=1}^{T}.
\end{equation}
This separation is the formal boundary for RQ1 and RQ4: $H_u^{(T)}$ supports the evaluated attack on histories indexed by users, while $G^{(T)}$ is a campaign level history.

\paragraph{Attack objective.}
Let $\mathcal{H}$ denote the space of histories indexed by user. An inference attack is a function
\begin{equation}
f:\mathcal{H}\rightarrow\{0,1\},
\qquad
\hat{S}_u=f(H_u^{(T)}).
\end{equation}
For probabilistic attacks, define
\begin{equation}
s_f(u)=\Pr_f(S_u=1\mid H_u^{(T)}).
\end{equation}
An adaptive policy selects legal campaigns by
\begin{equation}
c_t=\pi(H^{(t-1)},\mathcal{C}_{\mathrm{legal}},B_t),
\qquad
\sum_{t=1}^{T}\mathrm{cost}(c_t)\le B_0.
\end{equation}
This notation connects the oracle to RQ3: $f$ determines inference, while $\pi$ determines the next legal query.

\paragraph{Bayesian baseline.}
For a campaign $c$, sensitive value $s\in\{0,1\}$, and observation tied to a user $y$, define
\begin{equation}
q_{c,s}(y)=\Pr(Y_{u,c}=y\mid S_u=s,c).
\end{equation}
Let $\pi_0=\Pr(S_u=1)$. Under conditional independence of campaign observations given $S_u$, the posterior log odds are
\begin{equation}
\log\frac{\Pr(S_u=1\mid H_u^{(T)})}{\Pr(S_u=0\mid H_u^{(T)})}
=
\log\frac{\pi_0}{1-\pi_0}
+
\sum_{t=1}^{T}
\log
\frac{q_{c_t,1}(Y_{u,c_t})}{q_{c_t,0}(Y_{u,c_t})}.
\end{equation}
This expression defines the Bayesian attack used in the evaluation. The supervised attacks in Section~\ref{sec:attacks} use the same histories but relax this conditional independence assumption.

\paragraph{Disclosure postprocessing.}
Fix the campaigns and the exposure and interaction trace. If disclosure policy $D_2$ is a randomized postprocessing of disclosure policy $D_1$, then the observations tied to users released by $D_2$ contain no more information about $S_u$ than those released by $D_1$ in the evaluated channel tied to users. Let $O_1$ and $O_2$ be the corresponding observations tied to users. By the postprocessing condition, $O_2$ is sampled from $O_1$ without reading $S_u$ or any additional event state. Thus $S_u\rightarrow O_1\rightarrow O_2$ is a Markov chain, and the data processing inequality gives $I(S_u;O_2)\le I(S_u;O_1)$. Aggregate reporting removes the $H_u^{(T)}$ input required by the base attack on histories indexed by user; attacks over repeated $G^{(T)}$ are a separate aggregate attacker model.

The next section instantiates $f$ and $\pi$ with Bayesian, supervised, positive and unlabeled, and adaptive attacks while keeping the same campaign space, disclosure policies, and cost variables.

\section{Attacks}
\label{sec:attacks}

The attacks vary along two axes: the inference rule $f$ and the campaign selection policy $\pi$. Let $H_T(u)=H_u^{(T)}$ denote the target history after $T$ campaigns, and let $C_L=\mathcal{C}_{\mathrm{legal}}$ denote the legal campaign space. All attacks use campaign metadata and outputs released under $D$; calibration data are available only in the benchmark or shadow population setting.

\paragraph{Attack surface.}
The attacker is the advertiser defined in the threat model. For each selected campaign $c_t$, the attacker knows the predicate $g_{c_t}$, where eligibility is evaluated through $g_{c_t}(X_u)$. The attacker also knows the creative or topic $a_{c_t}$, the cost variable $b_{c_t}$, the legal space $C_L$, and the remaining budget $B_t$. These records match advertiser side audit settings where campaign inputs and released outputs can reveal information about selected audiences \cite{Korolova2011PrivacyViolationsUsingMicrotargeted,Venkatadri2018PrivacyRisksFacebooks,Lam2023SociotechnicalAuditsBroadeningAlgorithm}.

The released observation depends on $D$. Under disclosure with visible identity, the attacker observes $Y_{u,c_t}$ when user $u$ appears in a disclosed interaction record. Under reporting only or mixed disclosure, the attacker may observe aggregate outputs $Y_{c_t}$ or histories $G^{(T)}$. Such outputs can guide later campaign choices, but inference about a target user depends on the target history available to $f$. Engagement that exposes identity is the channel that binds a person to the campaign that induced the observation \cite{Kieserman2026RecklessDesignsBrokenPromises}.

\paragraph{Bayesian oracle attack.}
The Bayesian attack instantiates $f$ as a posterior threshold rule. Its score is
\begin{equation}
s_{\mathrm{Bayes}}(u)=\Pr(S_u=1\mid H_T(u)).
\end{equation}
Given a confidence threshold $\tau$, the decision rule is
\begin{equation}
\hat{S}_u=\mathbb{1}\{s_{\mathrm{Bayes}}(u)\ge \tau\}.
\end{equation}
For campaign $c$, sensitive value $s\in\{0,1\}$, and observation $y\in\mathcal{Y}$, the attack estimates
\begin{equation}
q_{c,s}(y)=\Pr(Y_{u,c}=y\mid S_u=s,c).
\end{equation}
The value $q_{c,s}(y)$ is estimated from calibration users or simulator parameters. The posterior factorization is defined in Section~\ref{sec:oracle-model}; this section uses it as a baseline rather than repeating the derivation. The baseline is useful because each campaign contributes an explicit evidence term, which makes it possible to compare later learned attacks against a transparent probabilistic rule. This framing follows prior membership and attribute inference work that estimates hidden properties from observable system behavior \cite{Shokri2017MembershipInferenceAttacksagainst,Jayaraman2022AreAttributeInferenceAttacks}.

\paragraph{Supervised shadow attack.}
The supervised attack replaces the explicit likelihood model with a learned predictor. Define the feature map
\begin{equation}
\phi(H_T(u))\in\mathbb{R}^d.
\end{equation}
The learned score is
\begin{equation}
s_{\mathrm{sup}}(u)=h_{\omega}(\phi(H_T(u))).
\end{equation}
The vector $\phi(H_T(u))$ encodes disclosed interaction indicators. It can also encode interaction types, aggregate outputs, predicate metadata, and cost features. The classifier $h_{\omega}$ is trained on synthetic or shadow users with known $S_u$ and evaluated on held out target users. The benchmark uses logistic regression, random forests, and gradient boosted trees as classifier families.

This attack relaxes the conditional independence assumption used by the Bayesian baseline. A learned model can represent correlations among predicates when those correlations appear in the shadow data. It can also represent repeated observation patterns induced by delivery and disclosure. The input boundary remains the same: histories are created from selected campaigns and outputs released under $D$. Learned privacy attacks motivate comparing such predictors with explicit posterior rules \cite{Rigaki2023SurveyPrivacyAttacksMachine,Shokri2017MembershipInferenceAttacksagainst}.

\paragraph{Positive and unlabeled attack.}
Disclosure with visible identity makes an observed interaction positive evidence, while absence from a disclosed list is censored. Define
\begin{equation}
Z_{u,c}=\mathbb{1}\{Y_{u,c}\text{ exposes identity}\}.
\end{equation}
The closed world setting treats missing disclosed records as negative observations:
\begin{equation}
Z_{u,c}=0\quad\text{is treated as negative.}
\end{equation}
The open world setting treats the same value as unlabeled:
\begin{equation}
Z_{u,c}=0\quad\text{is treated as unlabeled.}
\end{equation}
The closed world variant tests the effect of complete observation over target disclosures. The open world variant tests the effect of censoring by delivery, attention, interaction, or disclosure. Comparing the two settings measures sensitivity to observation completeness, which is central when an advertiser sees visible engagements but not a complete absent event log \cite{Kieserman2026RecklessDesignsBrokenPromises,Jayaraman2022AreAttributeInferenceAttacks}.

\paragraph{Adaptive campaign selection.}
The adaptive attack chooses later campaigns after observing earlier outputs. Its policy is
\begin{equation}
c_t=\pi(H^{(t-1)},C_L,B_t).
\end{equation}
For $p\in(0,1)$, define binary entropy as
\begin{equation}
H_b(p)=-p\log p-(1-p)\log(1-p).
\end{equation}
Let $s_{t-1}(u)$ be the current posterior score for target $u$. The expected information gain of candidate campaign $c$ is
\begin{equation}
\mathrm{EIG}(c\mid H^{(t-1)})
=
H_b(s_{t-1}(u))-\mathbb{E}_{Y}\left[H_b(s_t(u;Y))\right],
\end{equation}
where the expectation is over $Y\sim P(\cdot\mid c,H^{(t-1)})$. The normalized by cost rule is
\begin{equation}
c_t
=
\arg\max_{\substack{c\in C_L\\ \mathrm{cost}(c)\le B_t}}
\frac{\mathrm{EIG}(c\mid H^{(t-1)})}{\mathrm{cost}(c)+\lambda}.
\end{equation}
Here $\lambda>0$ is a stabilizer for small costs. This rule selects legal campaigns by expected uncertainty reduction per unit cost. Nonadaptive attacks fix the campaign sequence before observations, while adaptive attacks treat the advertising interface as an interactive query surface. Audit work motivates this view because later queries can depend on earlier platform outputs \cite{Lam2023SociotechnicalAuditsBroadeningAlgorithm}.

Aggregate reports can enter the expectation through $Y_c$ or $G^{(T)}$ when the disclosure policy releases campaign level outputs. These reports can update base rate estimates or change the estimated value of future queries when no observation indexed by user is released. The final decision for a target still uses the inference rule $f$ on the available target history. This separation preserves the distinction between histories indexed by user and aggregate histories established by the oracle model.

\paragraph{Stopping rules and baselines.}
For a confidence threshold $\tau$, define the first threshold crossing time as
\begin{equation}
T_{\tau}(u)=\min\{t:\max(s_t(u),1-s_t(u))\ge \tau\}.
\end{equation}
The campaign cost incurred by that time is
\begin{equation}
C_{\tau}(u)=\sum_{t=1}^{T_{\tau}(u)}\mathrm{cost}(c_t).
\end{equation}
A run without threshold crossing before budget exhaustion is recorded as a no threshold run. This convention separates failed confidence attainment from late success and lets the evaluation compare accuracy, confidence, and cost under the same trace.

The benchmark evaluates random legal campaign selection, nonadaptive informative campaigns, Bayesian posterior inference, and supervised shadow inference. It also evaluates positive and unlabeled inference, greedy expected information gain selection, and normalized by cost adaptive selection. These baselines instantiate both axes of the attack definition: $f$ determines how $H_T(u)$ is mapped to $\hat{S}_u$, while $\pi$ determines which legal campaign is queried next.

The next section constructs the benchmark users and campaign predicates. It then defines event models, disclosure policies, and calibration splits needed to instantiate these attacks.

\section{Benchmark and Evaluation Setup}
\label{sec:benchmark}
\label{sec:evaluation-setup}

\paragraph{Benchmark goals.}
The benchmark instantiates the oracle model and attacks as a reproducible artifact. It generates a user population $U$, observable features $X_u$, hidden labels $S_u$, legal campaigns $C_L$, exposure events $E_{u,c}$, interaction events $R_{u,c}$, disclosed observations $Y_{u,c}$, user histories $H_T(u)$, aggregate histories $G^{(T)}$, and campaign costs. The same interface supports the RQ2 sweeps over campaign count, interaction rate, predicate granularity, minimum audience size, and disclosure policy. Run metadata record the population size, campaign count, topic library size, target label, prevalence, random seeds, and exported evidence needed to reproduce a run.

\paragraph{Synthetic users and campaign semantics.}
Each generated user has platform side features $X_u$ and a hidden benchmark label $S_u$. The feature vector contains demographic buckets, household-status proxies, interest proxies, activity level, comment propensity, and reaction propensity. The label generator uses
\begin{equation}
\Pr(S_u=1\mid X_u)=\sigma(\gamma_0+\gamma^\top \phi(X_u)).
\end{equation}
Here $\sigma$ is the logistic link and $\phi(X_u)$ is the generator feature map. The evaluator observes $S_u$ for scoring, while the attacker observes only campaign histories and permitted calibration data. The artifact can use an LLM to improve campaign semantics under a fixed schema. This layer proposes topic names, short ad texts, proxy feature mappings, sensitive risk tags, and relative response tendencies. The simulator clips, normalizes, audits, and logs those values before using them as response priors. It still generates $S_u$, $E_{u,c}$, $R_{u,c}$, $Y_{u,c}$, attack labels, and metrics. A deterministic fallback topic library keeps the benchmark runnable without external API access.

\paragraph{Campaigns and events.}
Each campaign is $c=(g_c,a_c,b_c)$. The predicate $g_c$ is a Boolean expression over generated features, the topic $a_c$ is drawn from the topic library, and $b_c$ is a normalized campaign cost. The induced audience
\begin{equation}
A_c=\{u\in U:g_c(X_u)=1\}
\end{equation}
must satisfy $|A_c|\ge m$. Predicate length controls targeting granularity, subject to the same legality and audience-size rules at every length. Eligibility increases exposure probability but does not determine exposure. The simulator first samples $E_{u,c}$ and then samples $R_{u,c}$ conditional on exposure:
\begin{equation}
\Pr(R_{u,c}=1\mid E_{u,c}=1)=p_R(u,c;X_u,S_u,a_c,\theta_R).
\end{equation}
The parameter $\theta_R$ contains the base interaction rate, topic relevance weights, response noise, and feature correlation controls. The simulator enforces $\Pr(R_{u,c}=1\mid E_{u,c}=0)=0$. Tracking strength is a benchmark knob that changes predicate precision and correlation strength; it is not a live platform measurement.

\paragraph{Disclosure and splits.}
The benchmark implements disclosure with visible identity, type filtered disclosure, aggregate reporting, $k$-threshold disclosure, and randomized identity disclosure. Identity exposure and type filtered policies produce $H_T(u)$. Thresholded and randomized policies may produce partial user histories. Aggregate reporting produces $G^{(T)}$ rather than indexed by user histories. Defense experiments reuse the same event traces when possible and vary only the disclosure map. Generated users are split into calibration, training, and held out sets. Calibration estimates likelihood terms such as $q_{c,s}(y)$. Training supports supervised shadow attacks. Held out users support attack evaluation under fixed campaign and disclosure definitions.

\begin{table}[t]
\centering
\small
\caption{Benchmark parameters recorded by the artifact.}
\label{tab:benchmark parameters}
\resizebox{\linewidth}{!}{%
\begin{tabular}{ll}
\hline
Parameter & Role \\
\hline
Population size & Number of generated users in $U$ \\
Campaign count & Number of candidate campaigns and sweep values \\
Topic library size & Number of topics available to $a_c$ \\
Target label & Hidden benchmark label $S_u$ \\
Target prevalence & Generator setting recorded in metadata \\
Campaign count grid & Values of $T$ used by campaign sweeps \\
Interaction rate grid & Base response probabilities for sensitivity runs \\
Disclosure policy & Map from events to observations visible to advertisers \\
Random seed & Reproducibility control for generation and sampling \\
\hline
\end{tabular}}
\end{table}

\paragraph{Attack baselines and metrics.}
The evaluation maps directly to RQ1 through RQ4. It measures whether $H_T(u)$ contains signal about $S_u$ beyond the prior, how the signal changes across benchmark knobs, whether adaptive selection reduces campaigns or cost to confidence, and how disclosure policies change the oracle input under a common event trace. All baselines use the same campaign library, split assignment, disclosure policy, and held out users. The baseline set includes prior baseline scoring, random legal selection, nonadaptive informative selection, Bayesian posterior inference, supervised shadow inference, positive and unlabeled inference, greedy expected information gain selection, and normalized by cost adaptive selection. Microtargeting and nanotargeting results motivate comparing legal predicate choice against prevalence and random legal selection \cite{Korolova2011PrivacyViolationsUsingMicrotargeted,GonzlezCabaas2021UniqueFacebookFormulationEvidence}.

\begin{table}[t]
\centering
\small
\caption{Evaluation metrics for inference quality, posterior confidence, and attack cost.}
\label{tab:evaluation metrics}
\resizebox{\linewidth}{!}{%
\begin{tabular}{lll}
\hline
Metric & Definition & Used for \\
\hline
$\mathrm{AUC}$ & Area under the ROC curve over held out users & Ranking quality \\
Average precision & Area under the precision recall curve & Imbalanced labels \\
Balanced accuracy & Mean of true positive rate and true negative rate & Thresholded inference \\
Precision, recall & Positive predictive value and true positive rate & Error profile \\
Posterior confidence & $\max(s(u),1-s(u))$ & Score certainty \\
$T_{\tau}(u)$ & First campaign index reaching confidence $\tau$ & Campaign efficiency \\
$C_{\tau}(u)$ & Cost accumulated through $T_{\tau}(u)$ & Cost efficiency \\
Threshold crossing rate & Fraction of users reaching $\tau$ & Stopping behavior \\
\hline
\end{tabular}}
\end{table}

\paragraph{Sweeps and validity checks.}
The campaign count sweep evaluates attacks after increasing prefixes of the campaign sequence. The interaction rate sweep changes the base response probability. The predicate granularity and minimum audience sweeps vary predicate length and $m$. The tracking strength sweep changes predicate precision and correlation strength in the generator \cite{Englehardt2016OnlineTracking1million}. The disclosure sweep runs common exposure and interaction traces through different maps $D$, and the cost threshold sweep varies $\tau$ and budget settings. Before interpreting results, the artifact checks split separation, common event traces for defense comparisons, topic library schema validity, the boundary between generated campaign semantics and generated by the simulator ground truth, likelihood sanity for $q_{c,s}(y)$, and negative controls from prior baseline and random selection baselines.

\section{Evaluation Results}
\label{sec:evaluation results}

\paragraph{Final run.}
This section reports the final benchmark results for the oracle defined in Sections~\ref{sec:oracle-model} and~\ref{sec:benchmark}. The final run used four automatically generated topic variants, seven simulator seeds, and two interaction settings, for $56$ total runs. Each run used $8000$ synthetic users and $160$ campaigns. The target label was \texttt{health\_interest}. Generated topic libraries influenced only campaign semantics and response priors. Labels $S_u$, exposure events $E_{u,c}$, interaction events $R_{u,c}$, disclosed observations $Y_{u,c}$, attack labels, and metrics were generated by the simulator.

The results answer RQ1 through RQ4. Repeated campaigns with identity exposure increase the evidence available in user histories $H_T(u)$, answering RQ1. The two interaction settings and campaign count sweeps measure how the channel changes with interaction density and repeated queries, answering RQ2. The confidence and cost curves evaluate the stopping behavior used by adaptive policies, answering RQ3. The disclosure policy comparison evaluates platform side defenses, answering RQ4.

\paragraph{Attack performance.}
Table~\ref{tab:main-attack-results} summarizes attack performance at the maximum campaign count. At $160$ campaigns, the Bayesian and supervised attacks reach about $0.64$ AUC in the main setting and about $0.65$ AUC in the higher interaction setting. Bayesian AUPRC follows the same pattern. The benchmark records about $1.04$ visible observations per evaluation user in the main setting and about $1.33$ in the higher interaction setting. These values show that repeated campaigns produce usable signal, but the signal remains noisy. The result matches the model in Section~\ref{sec:oracle-model}: campaign histories compose evidence, while delivery, interaction, and disclosure noise constrain recovery of $S_u$.

\begin{table*}[t]
\centering
\small
\caption{Main inference results at $160$ campaigns. Values are means across $28$ runs per setting, with standard deviations in parentheses. The main setting is the conservative configuration. The higher interaction setting increases event density while keeping the same population, campaign, and disclosure interface.}
\label{tab:main-attack-results}
\resizebox{0.96\textwidth}{!}{%
\begin{tabular}{lccccc}
\hline
Setting & Bayesian AUC & Supervised AUC & Bayesian AUPRC & Visible obs./user & $\Pr[\max(s,1-s)\ge 0.70]$ \\
\hline
Main & $0.636\,(0.012)$ & $0.636\,(0.010)$ & $0.642\,(0.016)$ & $1.042\,(0.050)$ & $0.172\,(0.019)$ \\
Higher interaction & $0.654\,(0.013)$ & $0.653\,(0.013)$ & $0.659\,(0.017)$ & $1.334\,(0.068)$ & $0.298\,(0.158)$ \\
\hline
\end{tabular}}
\end{table*}

The bounded strength is important for interpretation. The evaluated attacks are stronger than prior baseline scoring, but the oracle derives its signal from repeated observations under legal campaign predicates. This supports comparing oracle based inference against prevalence and imputation baseline baselines, rather than treating any ranking gain as direct disclosure \cite{Jayaraman2022AreAttributeInferenceAttacks}.

\begin{figure*}[t]
\centering
\subfloat[Main: AUC increases with the number of queried campaigns.\label{fig:attack-main}]{\includegraphics[width=0.48\textwidth]{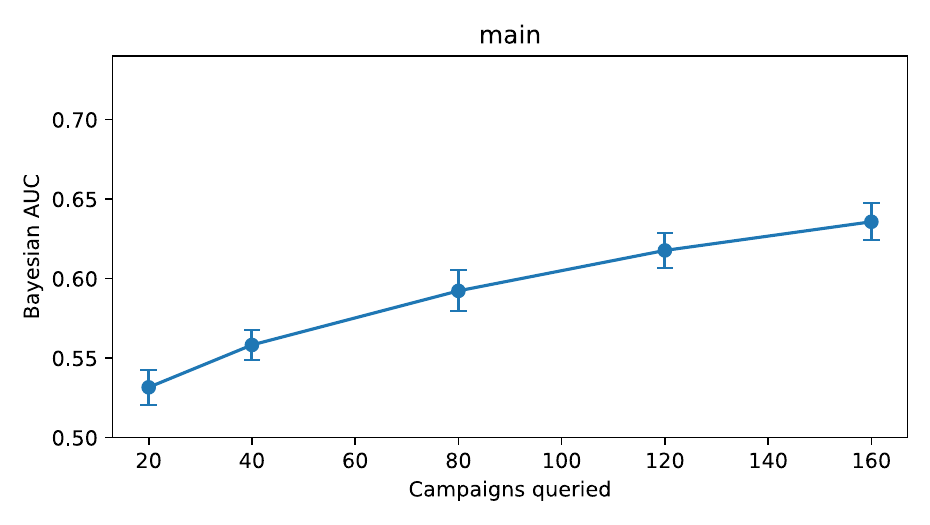}}
\hfill
\subfloat[Higher interaction: the same trend appears with a larger signal.\label{fig:attack-high}]{\includegraphics[width=0.48\textwidth]{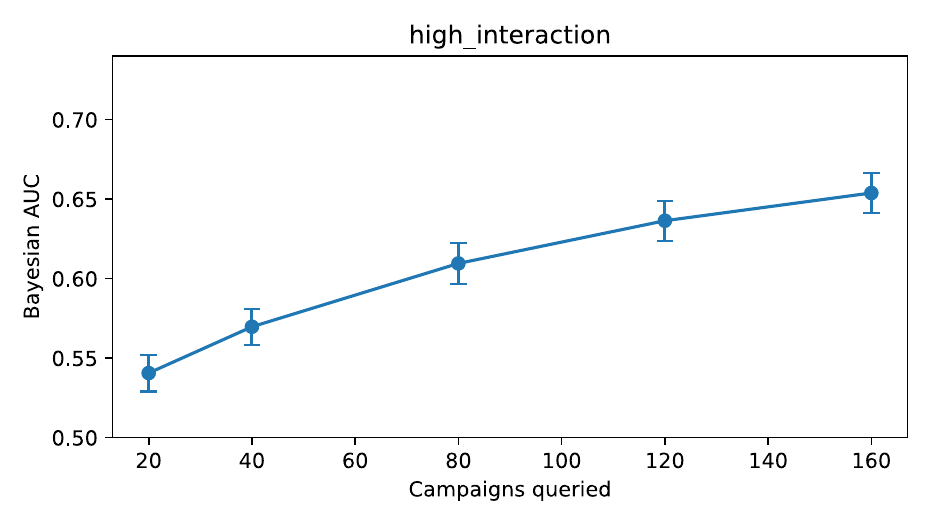}}
\vspace{0.5em}
\subfloat[Main: high confidence decisions remain limited.\label{fig:cost-main}]{\includegraphics[width=0.48\textwidth]{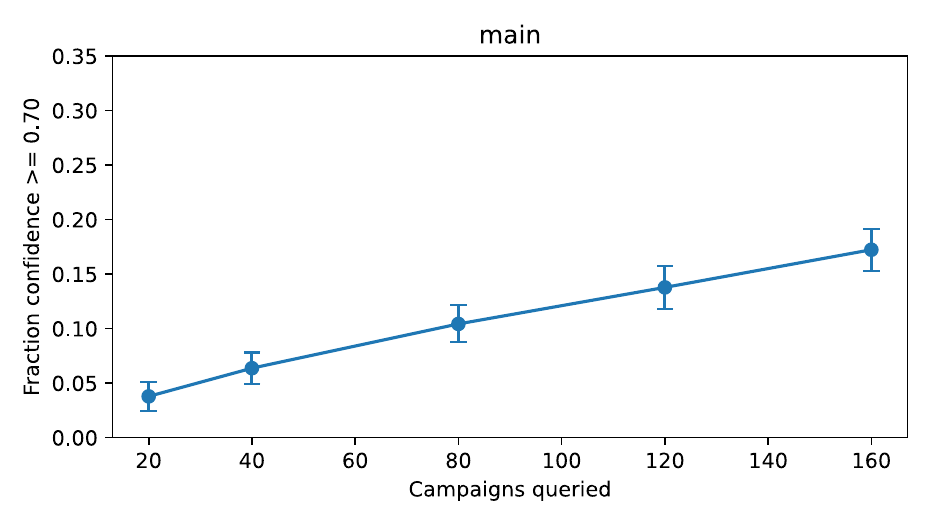}}
\hfill
\subfloat[Higher interaction: more observations increase confidence crossings.\label{fig:cost-high}]{\includegraphics[width=0.48\textwidth]{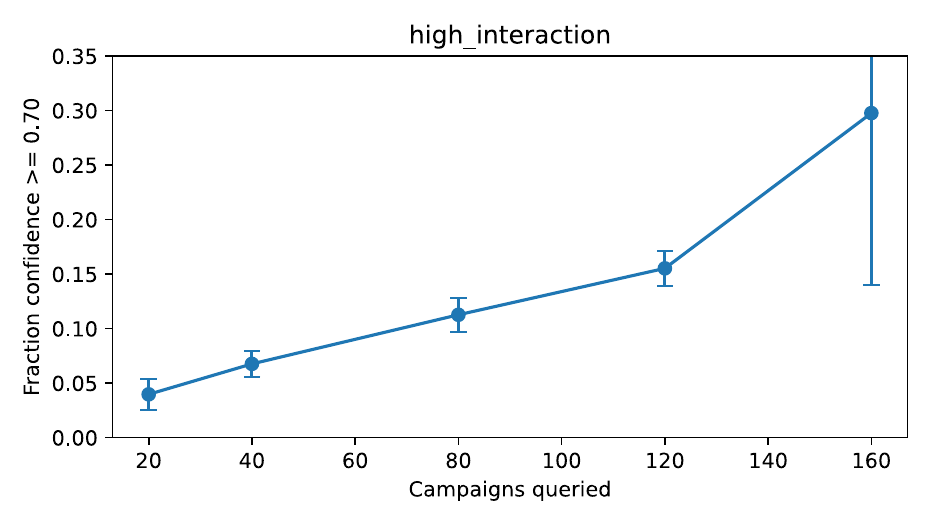}}
\caption{\textbf{Campaign count and confidence sweeps.} The top row shows the inference signal as campaigns accumulate. Bayesian AUC rises from $0.532$ to $0.636$ in the main setting and from $0.541$ to $0.654$ in the higher interaction setting. The bottom row reports the fraction of evaluated users whose posterior confidence crosses $0.70$. The same monotone pattern appears, but high confidence decisions remain a minority of runs in the main setting. Together, the four panels show that the oracle is compositional: repeated campaigns with identity exposure add evidence, while interaction noise keeps the attack bounded.}
\label{fig:campaign and confidence}
\end{figure*}

\paragraph{Disclosure policies.}
Table~\ref{tab:disclosure-results-final} shows that the disclosure map $D$ is the strongest control in both settings. Disclosure with visible identity gives the evaluated attacks a history indexed by user and yields Bayesian AUC $0.636$ in the main setting and $0.654$ in the higher interaction setting. Aggregate reporting remains at prior ranking with AUC $0.500$ and releases no observations tied to users in the base oracle. This matches the aggregate history definition and the disclosure postprocessing argument in Section~\ref{sec:oracle-model}: the evaluated attack on histories indexed by user needs an input history $H_T(u)$.

\begin{table*}[t]
\centering
\small
\caption{Disclosure policy effects under the final run. Values are mean Bayesian AUC across runs, with standard deviations in parentheses. Aggregate reporting remains at chance level ranking in both settings. Type filtering and randomized identity disclosure reduce the signal, while the evaluated threshold setting remains close to disclosure with visible identity.}
\label{tab:disclosure-results-final}
\resizebox{0.96\textwidth}{!}{%
\begin{tabular}{lcccccc}
\hline
Setting & Identity exposure & Type filtered & Randomized & $k$-threshold & Aggregate reports & Identity to aggregate drop \\
\hline
Main & $0.636\,(0.012)$ & $0.584\,(0.012)$ & $0.592\,(0.010)$ & $0.636\,(0.012)$ & $0.500\,(0.000)$ & $0.136\,(0.012)$ \\
Higher interaction & $0.654\,(0.013)$ & $0.601\,(0.012)$ & $0.609\,(0.013)$ & $0.654\,(0.013)$ & $0.500\,(0.000)$ & $0.154\,(0.013)$ \\
\hline
\end{tabular}}
\end{table*}

Type filtering reduces Bayesian AUC to $0.584$ in the main setting and $0.601$ in the higher interaction setting. Randomized identity disclosure weakens the oracle to $0.592$ and $0.609$. Thresholded disclosure remains close to disclosure with visible identity in both configurations, indicating that the evaluated threshold does not bind strongly under these event distributions. These results support evaluating defenses as transformations of $D$, not only as written platform policies. Thresholds, randomized release, and aggregate reporting are standard mechanisms for reducing individual disclosure from released records or statistics \cite{Sweeney2002kAnonymityModelProtecting,Warner1965RandomizedResponseSurveyTechnique,Dwork2006CalibratingNoiseSensitivityPrivate}.

\begin{figure*}[t]
\centering
\subfloat[Main: disclosure policy AUC.\label{fig:def-main}]{\includegraphics[width=0.48\textwidth]{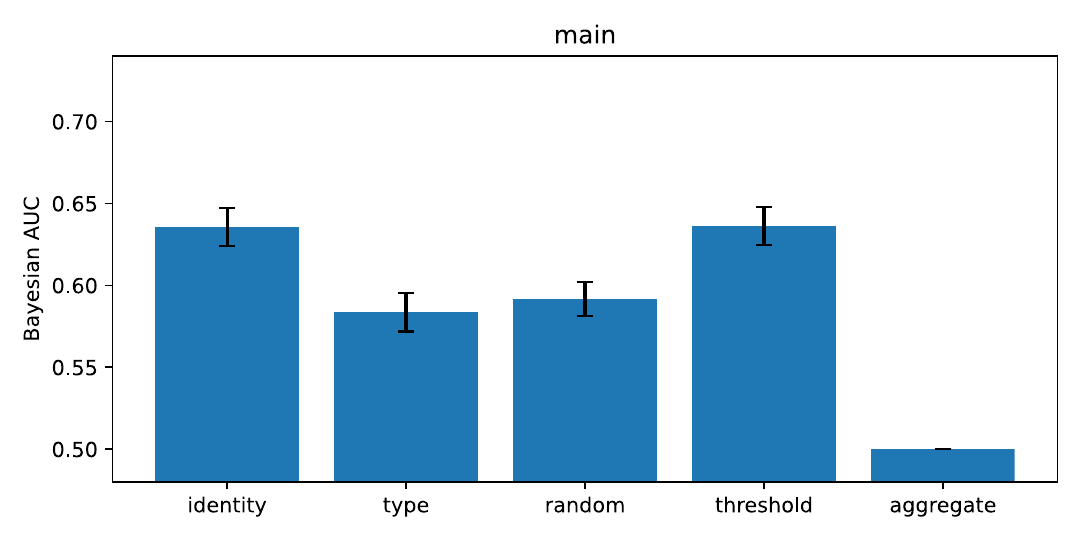}}
\hfill
\subfloat[Higher interaction: disclosure policy AUC.\label{fig:def-high}]{\includegraphics[width=0.48\textwidth]{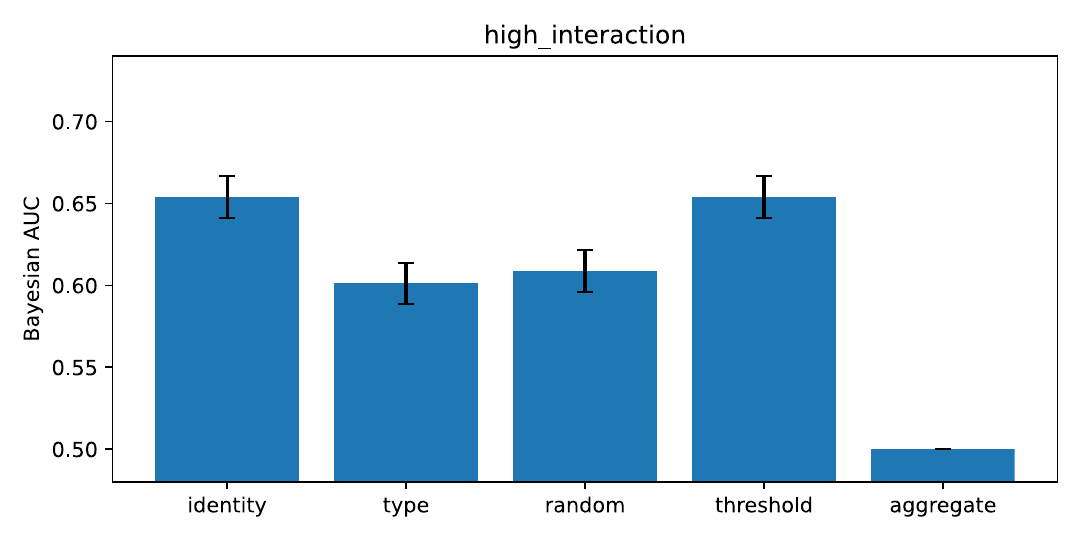}}
\vspace{0.5em}
\subfloat[Main: AUC reduction relative to disclosure with visible identity.\label{fig:def-drop-main}]{\includegraphics[width=0.48\textwidth]{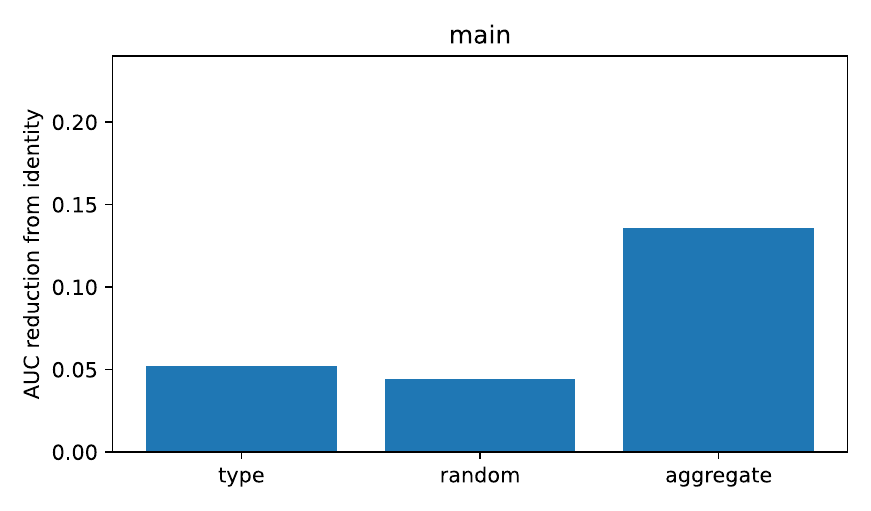}}
\hfill
\subfloat[Higher interaction: AUC reduction relative to disclosure with visible identity.\label{fig:def-drop-high}]{\includegraphics[width=0.48\textwidth]{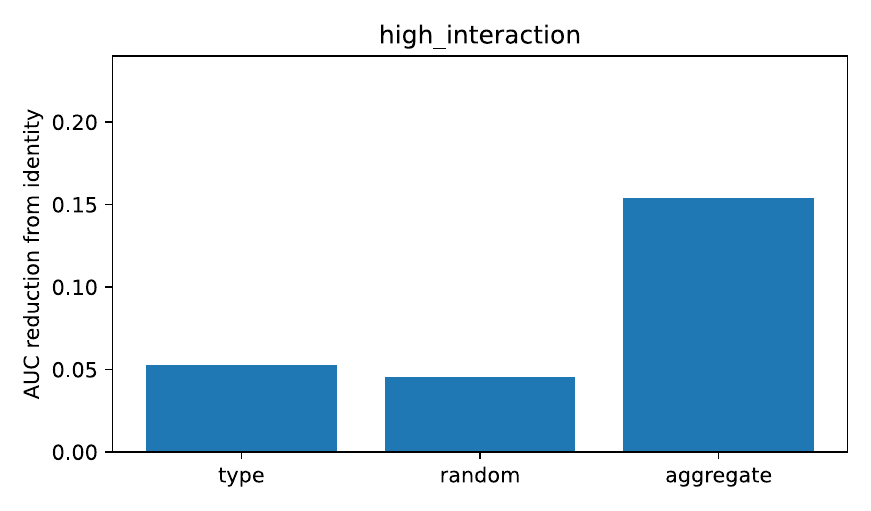}}
\caption{\textbf{Disclosure policy is the dominant platform control.} The top row shows Bayesian AUC under identity exposure, type filtering, randomized identity release, thresholding, and aggregate disclosure. The bottom row shows the corresponding AUC reduction relative to disclosure with visible identity. Aggregate reporting returns the evaluated oracle input tied to users to chance level ranking in both settings. Type filtering and randomized identity disclosure reduce, but do not remove, the signal. The evaluated threshold setting stays near the identity exposure baseline because the release threshold does not bind strongly for the simulated event rates.}
\label{fig:disclosure-policies}
\end{figure*}

\begin{figure*}[t]
\centering
\subfloat[Interaction setting sensitivity: higher interaction increases AUC.\label{fig:setting-sens}]{\includegraphics[width=0.31\textwidth]{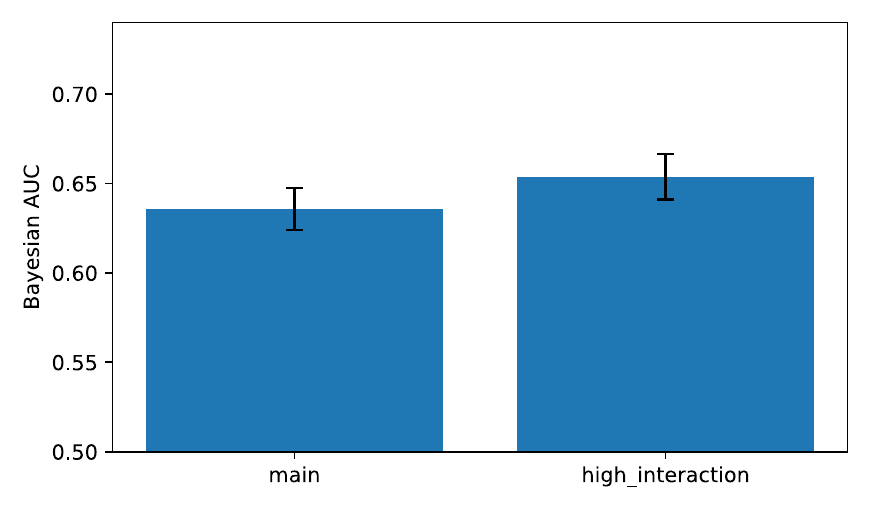}}
\hfill
\subfloat[Main: topic library robustness.\label{fig:topic-main}]{\includegraphics[width=0.31\textwidth]{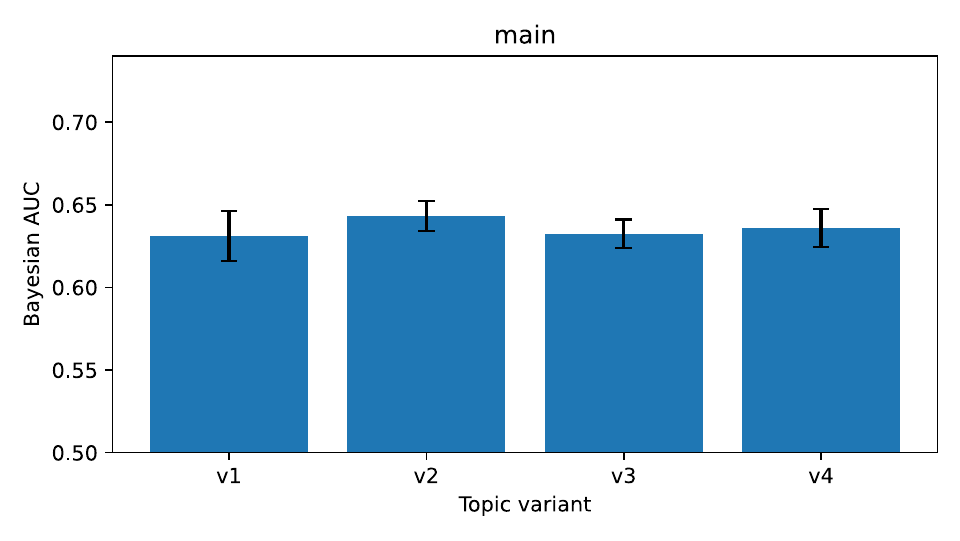}}
\hfill
\subfloat[Higher interaction: topic library robustness.\label{fig:topic-high}]{\includegraphics[width=0.31\textwidth]{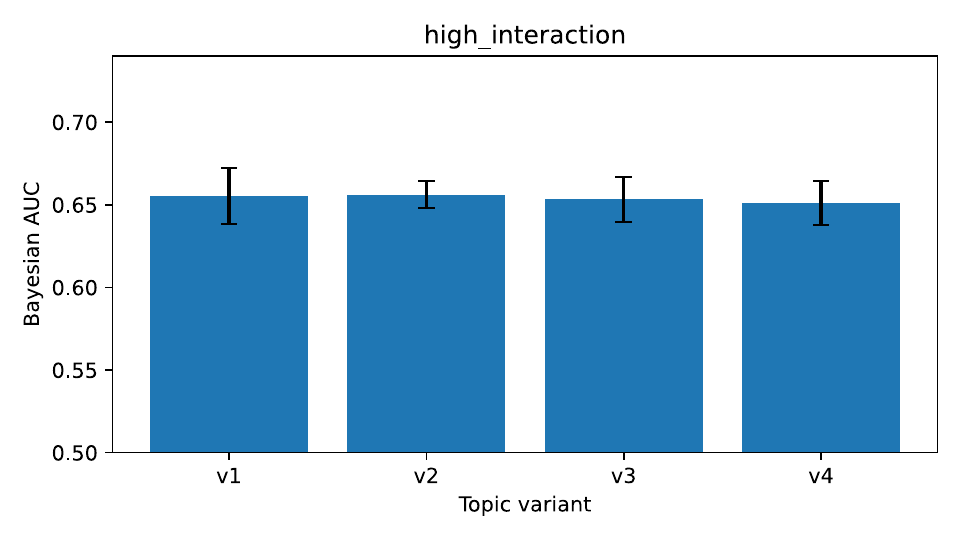}}
\vspace{0.5em}
\subfloat[Main: predicate granularity sweep.\label{fig:pred-main}]{\includegraphics[width=0.31\textwidth]{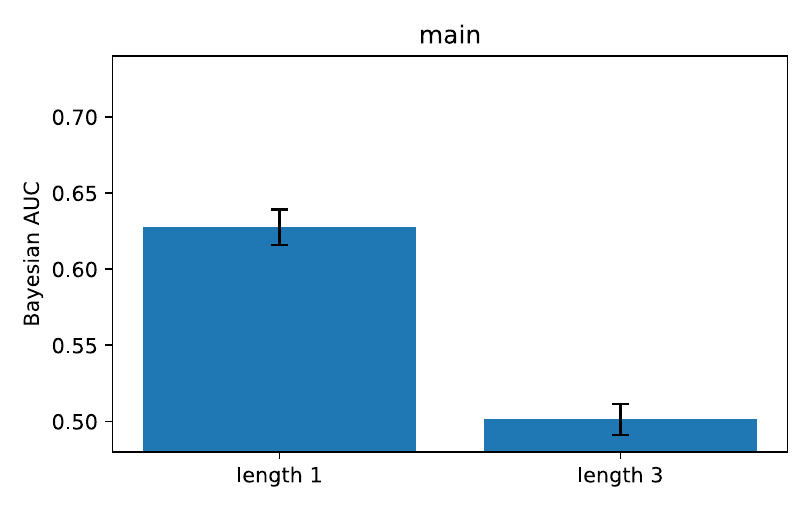}}
\hfill
\subfloat[Higher interaction: predicate granularity sweep.\label{fig:pred-high}]{\includegraphics[width=0.31\textwidth]{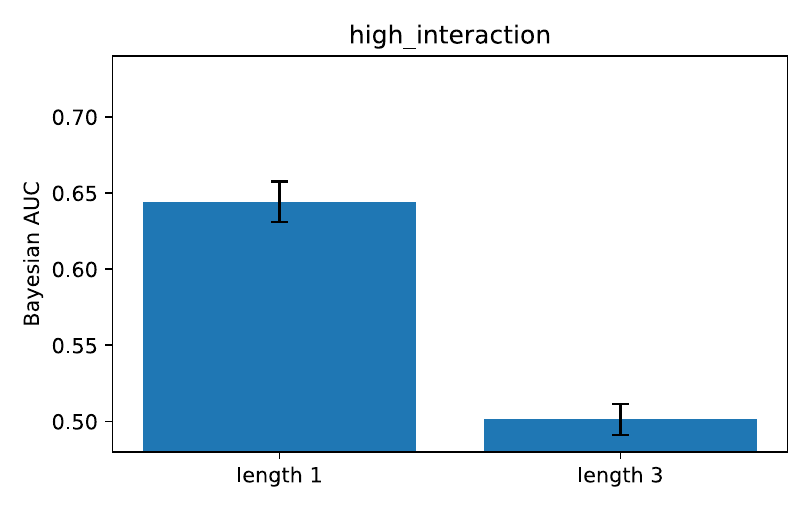}}
\hfill
\subfloat[Main: visibility rate heatmap by topic risk and predicate length.\label{fig:heat-main}]{\includegraphics[width=0.31\textwidth]{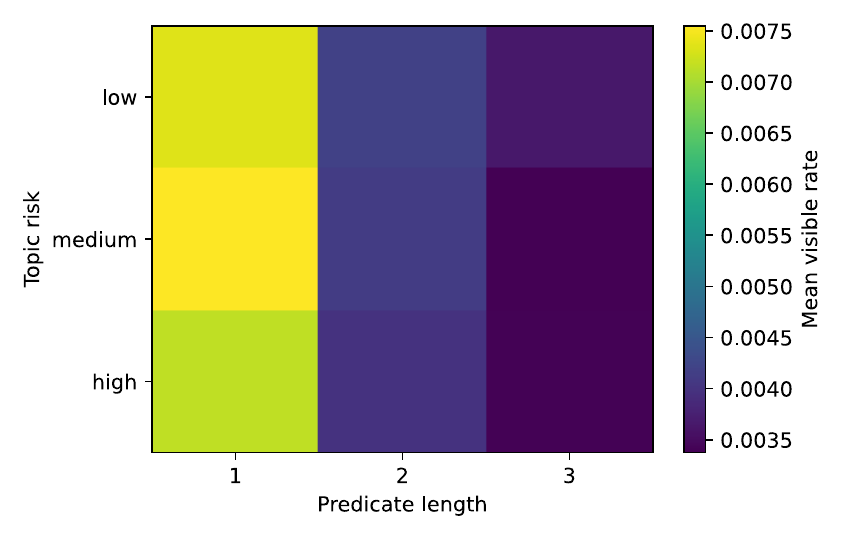}}
\caption{\textbf{Robustness and diagnostic checks.} The top row summarizes interaction setting sensitivity and topic library robustness. The higher interaction setting increases the measured signal, but the same qualitative pattern remains: histories with identity exposure support bounded inference and aggregate disclosure removes the evaluated oracle input tied to users. The bottom row reports predicate granularity diagnostics. Predicate length $1$ retains measurable signal, while predicate length $3$ is close to prior ranking because narrower predicates reduce exposure and interaction opportunities under the minimum audience constraint. The heatmap gives the corresponding visibility pattern in the main setting.}
\label{fig:robustness-diagnostics}
\end{figure*}

\paragraph{Topic library robustness and interaction setting sensitivity.}
Figure~\ref{fig:robustness-diagnostics} shows that the main result is stable across generated topic libraries and across interaction settings. The four variants use different topic library seeds and temperatures. Bayesian AUC stays in a narrow band around the setting mean. The higher interaction setting increases AUC from $0.636$ to $0.654$ and increases visible observations per user from $1.04$ to $1.33$. This direction matches the oracle model: more visible evidence strengthens inference, while aggregate disclosure still removes the evaluated input tied to users.

The robustness result addresses a concern specific to the benchmark. Generated topics are used to improve campaign semantics and response priors, while labels and event outcomes come from the simulator. Stable behavior across topic variants indicates that the result is a property of the oracle channel grounded in the simulator: campaign predicates, sampled exposure, sampled interaction, and the disclosure map.

\paragraph{Cost and predicate granularity.}
Table~\ref{tab:cost-granularity-summary} summarizes two secondary checks. At confidence threshold $\tau=0.70$, the threshold crossing rate is $0.172$ in the main setting and $0.298$ in the higher interaction setting. This means that high confidence inference is reached for a minority of target runs under the campaign limit in the main setting, even though ranking performance is above chance. The cost to confidence result therefore constrains the attack interpretation: the ranking signal is measurable, but high posterior confidence is not routine for every target.

\begin{table}[t]
\centering
\small
\caption{Cost and predicate granularity summary. Values are means across $28$ runs per setting, with standard deviations in parentheses. Predicate length $1$ retains measurable signal, while predicate length $3$ is near prior ranking in both settings.}
\label{tab:cost-granularity-summary}
\resizebox{\linewidth}{!}{%
\begin{tabular}{lccc}
\hline
Setting & $\Pr[\max(s,1-s)\ge0.70]$ & Length $1$ AUC & Length $3$ AUC \\
\hline
Main & $0.172\,(0.019)$ & $0.628\,(0.012)$ & $0.501\,(0.010)$ \\
Higher interaction & $0.298\,(0.158)$ & $0.644\,(0.013)$ & $0.501\,(0.010)$ \\
\hline
\end{tabular}}
\end{table}

The predicate granularity check shows that narrower predicates do not automatically improve inference. Predicate length $1$ gives Bayesian AUC $0.628$ in the main setting and $0.644$ in the higher interaction setting, while predicate length $3$ stays near $0.501$. Longer predicates narrow the eligible audience, but they also reduce opportunities for exposure and interaction under minimum audience constraints. This result keeps the threat model tied to the actual query surface: highly specific legal predicates remain relevant to privacy, but their value depends on whether the disclosure channel supplies enough observations indexed by user.

\paragraph{Summary.}
The final run connects the four RQs to the measured oracle behavior. For RQ1, repeated campaigns with identity exposure create a measurable and stable but bounded inference signal. For RQ2, the signal grows with campaign count and interaction density. For RQ3, confidence crossing remains limited in the main setting and increases in the higher interaction setting. For RQ4, disclosure policy controls the channel tied to users: type filtering and randomized disclosure reduce the signal, and aggregate reporting remains at prior ranking in both settings. The measured oracle is stable across topic variants, and aggregate reporting removes the evaluated input $H_T(u)$.

\section{Discussion}
\label{sec:discussion}

\paragraph{What the benchmark establishes.}
The benchmark gives a formal and empirical account of interactive targeted ads with identity exposure as noisy attribute inference oracles. The model fixes the roles of campaign predicates, delivery, interaction, and disclosure. The artifact then instantiates those roles with known labels, legal campaigns, event traces, and release policies. The final results answer RQ1 and RQ2 by showing that repeated campaigns with identity exposure create measurable signal in $H_T(u)$ across two interaction settings. They answer RQ4 by showing that the disclosure map $D$ controls whether the evaluated attacks receive a history indexed by user.

\paragraph{Interpreting bounded attack strength.}
AUC around $0.64$ in the main setting and $0.65$ in the higher interaction setting indicates stable inference signal under noisy delivery and interaction. This distinction matters for platform design. The oracle becomes meaningful through repeated observations, and the confidence curves quantify how often repeated campaigns reach high posterior confidence. The result supports the central model: interactive targeted ads form a noisy oracle whose risk depends on campaign repetition and disclosure design. The bounded signal also makes the defense analysis sharper. A release policy can reduce or remove the history indexed by user used by the evaluated attacks, even when the underlying event process still produces engagements.

\paragraph{Calibration and interface validation.}
The evaluation uses synthetic users calibrated from public data because the oracle channel requires known sensitive labels for scoring. This design isolates the channel and supports controlled sweeps over campaign count, predicate granularity, interaction rate, tracking strength, and disclosure policy. Real platform validation fits this structure as an interface check. Controlled or consented accounts can confirm whether a platform exposes the interaction records represented by $Y_{u,c}$. Redacted or aggregate evidence can document disclosure behavior while retaining the benchmark as the place where attribute inference is measured.

\paragraph{Automatic topic generation.}
The automatic topic layer improves semantic realism by producing campaign topics, proxy feature mappings, sensitive risk tags, and response prior suggestions. Generated topic libraries influence only campaign semantics and response priors. Labels $S_u$, exposure events $E_{u,c}$, interaction events $R_{u,c}$, disclosed observations $Y_{u,c}$, attack labels, and metrics are generated by the simulator. The final run uses four topic variants, seven simulator seeds, and two interaction settings, for $56$ total runs. Bayesian and supervised AUC reach $0.636$ in the main setting and $0.654$ or $0.653$ in the higher interaction setting, and aggregate reporting stays at prior ranking in the evaluated base oracle. This supports the conclusion that the disclosure effect is stable across topic libraries and interaction rates.

\paragraph{Defense implications.}
The strongest implication concerns disclosure. Aggregate reporting removes the histories indexed by user used by the evaluated attacks. Type filtering reduces the signal. Randomized identity disclosure weakens the oracle. Thresholding depends on whether the threshold binds under the event distribution. These findings support evaluating defenses as transformations of $D$. Written targeting policies matter, but the released observation determines whether an advertiser can construct $H_T(u)$ for the base attack on histories indexed by user. This disclosure side view also frames interactive advertising as a platform governance problem rather than only a targeting policy problem. Advertising platforms, like broader AI marketplaces, must balance data utility, participant incentives, trust, auditability, and privacy preserving access to signals derived from users \cite{li2023web3ai}.

\paragraph{Extensions.}
The artifact supports extensions that preserve the same oracle interface. Public data calibration can replace synthetic marginals with demographic and survey marginals. Larger seed sweeps can tighten uncertainty intervals. Controlled validation can measure disclosure behavior on consented accounts. Browser and tracking strength experiments can be represented as parameterized changes to predicate precision and feature correlation. Inference from aggregate reports can be evaluated as a separate attacker model over $G^{(T)}$. These extensions keep the contribution centered on the same object: a model, benchmark, and defense analysis for leakage through records tied to users created by interactive targeted advertising disclosure.

\section{Conclusion}
\label{sec:conclusion}

Interactive targeted ads combine predicates chosen by advertisers with interactions that expose identity. This combination creates an observation channel tied to users. We model that channel as a noisy attribute inference oracle with separate stages for campaign predicates, exposure, interaction, and disclosure.

The benchmark gives this oracle an executable form. It uses synthetic users calibrated from public data, known benchmark labels, legal campaign predicates, sampled exposure and interaction events, and controlled disclosure maps. Automatic campaign semantics make the campaign library more realistic, while the simulator remains the source of labels, events, disclosed observations, and metrics.

Across four topic variants, seven simulator seeds, and two interaction settings, repeated campaigns with identity exposure produce measurable but bounded inference signal. At $160$ campaigns, Bayesian and supervised attacks reach about $0.64$ AUC in the main setting and about $0.65$ AUC in the higher interaction setting. Disclosure policy is the dominant control. Aggregate reporting removes the evaluated oracle input tied to users. Type filtering and randomized disclosure reduce the released signal.

Disclosure design at the platform therefore matters. Defenses should be evaluated at the disclosure map $D$ and reporting interface, not only at targeting policy text. The result is a model, benchmark, and defense evaluation method for reasoning about privacy leakage from interactive targeted advertising systems.

\bibliographystyle{IEEEtran}
\bibliography{d_aggregate_references,interactive_ad_extra_refs}

\end{document}